\documentclass{article}

\usepackage{arxiv}
\usepackage{framed,multirow}

\usepackage{xcolor}
\usepackage{amsmath}
\newcommand{\argmax}[1]{\underset{#1}{\operatorname{arg}\,\operatorname{max}}\;}

\newcommand{\red}[1]{\textcolor{red}{#1}}
\newcommand{\one}[1]{\textbf{#1}}

\usepackage{amsmath}
\usepackage{subfigure}
\usepackage{amssymb}
\usepackage{multirow}
\usepackage[colorinlistoftodos]{todonotes}
\usepackage[breaklinks=true,colorlinks,bookmarks=false]{hyperref}

\usepackage{setspace}

\title{Self-Balanced R-CNN for Instance Segmentation}

\author{ \href{https://orcid.org/0000-0002-9316-595X}{\includegraphics[scale=0.06]{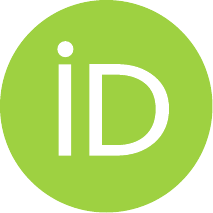}\hspace{1mm}Leonardo Rossi} \\
    Department of Engineering and Architecture\\
    University of Parma,\\
    Parco Area delle Scienze, 181/A, I-43124 Parma, Italy \\
    \texttt{leonardo.rossi@unipr.it} \\
    \And
    Akbar Karimi \\
    Department of Engineering and Architecture\\
    University of Parma,\\
    Parco Area delle Scienze, 181/A, I-43124 Parma, Italy \\
    \texttt{akbar.karimi@unipr.it} \\
    \And
    \href{https://orcid.org/0000-0002-1211-529X}{\includegraphics[scale=0.06]{images/orcid.pdf}\hspace{1mm}Andrea Prati} \\
    Department of Engineering and Architecture\\
    University of Parma,\\
    Parco Area delle Scienze, 181/A, I-43124 Parma, Italy \\
    \texttt{andrea.prati@unipr.it} \\
}

\begin{document}

\maketitle

\begin{abstract}
Current state-of-the-art two-stage models on instance segmentation task suffer from several types of imbalances.
In this paper, we address the Intersection over the Union (IoU) distribution imbalance of positive input Regions of Interest (RoIs) during the training of the second stage.
Our Self-Balanced R-CNN (SBR-CNN), an evolved version of the Hybrid Task Cascade (HTC) model, brings brand new loop mechanisms of bounding box and mask refinements.
With an improved Generic RoI Extraction (GRoIE), we also address the feature-level imbalance at the Feature Pyramid Network (FPN) level, originated by a non-uniform integration between low- and high-level features from the backbone layers.
In addition, the redesign of the architecture heads toward a fully convolutional approach with FCC further reduces the number of parameters and obtains more clues to the connection between the task to solve and the layers used.
Moreover, our SBR-CNN model shows the same or even better improvements if adopted in conjunction with other state-of-the-art models.
In fact, with a lightweight ResNet-50 as backbone, evaluated on COCO minival 2017 dataset, our model reaches 45.3\% and 41.5\% AP for object detection and instance segmentation, with 12 epochs and without extra tricks.
The code is available at \url{https://github.com/IMPLabUniPr/mmdetection/tree/sbr_cnn}.
\end{abstract}

\keywords{
	Object detection \and Instance segmentation \and Imbalance in R-CNN networks \and Two-stage deep learning architectures.
}



\section{Introduction}\label{sec:introduction}
Nowadays, instance segmentation is one of the most studied topics in the computer vision community, because it reflects one of the key problems for many of the existing applications where we have to deal with many heterogeneous objectives inside an image.
It offers, as output, the localization and segmentation of a number of instances not defined a priori, each of them belonging to a list of classes.
This task is important for several applications, including medical diagnostics \cite{chen2017dcan}, autonomous driving \cite{huang2019robust}, alarm systems \cite{tian2018eliminating}, agriculture optimization \cite{ge2019instance}, visual product search \cite{liu2015matching}, and many others.

Most of the recent models descend from the two-stage architecture called Mask R-CNN \cite{he2017mask}.
The first stage is devoted to the search of interesting regions independently from the class, while the second is used to perform classification, localization and segmentation on each of them.
This divide-and-conquer approach was first introduced in the ancestor network called Region-based CNN (R-CNN) \cite{girshick2014rich}, which has evolved in several successive architectures.
Although it achieved excellent results, several studies \cite{oksuz2020imbalance}, \cite{song2020revisiting}, \cite{zhang2020bridging} have recently discovered some of its critical issues which can limit its potentiality.
These issues have not been solved yet and several blocks of these architectures are still under-explored and far from optimized and well understood.

This paper approaches mainly two of the imbalance problems mentioned in \cite{oksuz2020imbalance}.
The first problem, called IoU Distribution Imbalance (IDI), arises when the positive Regions of Interest (RoIs) proposals provided by the RPN during the training of the detection and segmentation heads have an imbalanced distribution.
Due to some intrinsic problems of the anchor system, the number of available RoIs decreases exponentially with the increase of the IoU threshold, which leads the network to easily overfit to low quality proposals.
Our work extends the analysis on $R^3$-CNN, first introduced in \cite{rossi2021recursively}, to understand architectural limits and proposes advanced configurations in between and an architectural improvement for the segmentation head.

The second problem, called Feature Level Imbalance (FLI), arises when the features are selected from the Feature Pyramid Network (FPN) for their localization and segmentation.
As highlighted in \cite{oksuz2020imbalance}, the hierarchical structure of FPN (originally designed to provide multi-scale features) does not provide a good integration between low- and high-level features among different layers.
To address this problem, the classical approach is to balance the information before the FPN.
On the contrary, our work enhances the GRoIE \cite{rossi2020novel} architecture and puts forward a more effective solution, fusing information from all the FPN layers.

In addition, we address the common problem of the explosion of the number of parameters, due to the introduction of new components or expansion of existing ones (e.g. \cite{cai2018cascade}).
The increased complexity leads to an increase in the search space for optimization during the training, and, in turn, negatively impacts the generalization capability of the network.
Moreover, our empirical results support the intuition made by \cite{wu2020rethinking} about the connection between the task to solve and the utilized layers, extending their work toward a fully convolutional solution.

To summarize, this paper has the following main contributions:
\begin{itemize}
\item An extensive analysis of the IDI problem in the RPN generated proposals, which we treat with a single- and double-head loop architecture ($R^3$-CNN) between the detection head and the RoI extractor, and a brand-new internal loop for the segmentation head itself.
\item Redesign of the model heads (FCC) toward a fully convolutional approach, with empirical analysis that supports some architectural preferences depending on the task.
\item A better performing GRoIE model is proposed for extraction of RoIs in a two-stage instance segmentation and object detection architecture.
\item An exhaustive ablation study on all the components.
\item The proposal of SBR-CNN, a new architecture composed of $R^3$-CNN, FCC and GRoIE, which maintains its qualities if plugged into major state-of-the-art models.
\end{itemize}

The paper is organized as follows:
in Section \ref{Related Works}, state of the art related to the relevant topics is reported;
Section \ref{sec:sbrcnn} details each contribution of which the proposed SBR-CNN is composed;
Section \ref{experiments}, reports the extensive evaluation of the different architectural enhancements introduced, by conducting several ablation studies and a final experiment comparing SBR-CNN with some state-of-the-art models;
finally, Section \ref{conclusions} draws final conclusions about the proposed work and envisions possible future directions of research.


\section{Related Works}
\label{Related Works}

\noindent\textbf{Multi-stage Detection/Instance Segmentation}. 
Single-stage and two-stage architectures for object detection have been researched for several years.
For instance, YOLO network proposed in \cite{redmon2016you} optimizes localization and classification in one step and \cite{liu2016ssd} proposes a single-shot network which uses bounding box regression.
Since the single-stage architectures do not always provide acceptable performance and require a lot of memory in applications with thousands of classes, a region-based recognition method was proposed \cite{girshick2014rich}, where first part processes input images, while the second part processes bounding boxes found by the previous one.
This approach has been used in the Mask R-CNN architecture \cite{he2017mask}, obtained by adding a segmentation branch to the Faster R-CNN \cite{ren2015faster}.
This idea has been refined by several studies.
For instance, \cite{liu2020cbnet} provides a composite backbone network in a cascade fashion. 
The Cascade R-CNN architecture \cite{cai2018cascade} puts forward the utilization of multiple bounding box heads, which are sequentially connected, refining predictions at each stage.
In \cite{vu2019cascade, wang2019region, zhong2020cascade}, they introduced a similar cascade concept but applied to the RPN network.
In addition, the Hybrid Task Cascade (HTC) network \cite{chen2019hybrid}, by which this work is inspired, applies cascade operation on the mask head as well. 
Our work pushes in the same direction but changes the paradigm from cascade to loop, where the single neural network block is trained to perform more than one function by applying different conditioning in the input.

\noindent\textbf{IoU distribution imbalance}.
A two-stage network uses the first stage to produce a set of bounding box proposals for the following stage, filtering positive ones through a threshold applied to the IoU between them and the ground truth.
The IoU distribution imbalance problem is described as a skewed IoU distribution \cite{oksuz2020imbalance} that is seen in bounding boxes which are utilized in training and evaluation. 
In \cite{shrivastava2016training}, the authors propose a hard example mining algorithm to select the most significant RoIs to deal with background/foreground imbalance. Their work differs from ours because our primary goal is to balance the RoIs across the positive spectrum of the IoU. In \cite{pang2019libra}, the authors propose an IoU-balanced sampling method which mines the hard examples. The proposed sampling is performed on the results of the RPN which is not very optimized in producing high-quality RoIs as we will see. On the other hand, we apply the resampling on the detector itself, which increases the probability of returning more significant RoIs. 

After analyzing the sources of false positives and to reduce them, \cite{cheng2018revisiting} introduces an extra independent classification head to be attached to the original architecture. In \cite{zhu2021iou}, the authors propose a new IoU prediction branch which supports classification and localization. Instead of utilizing RPN for localization and IoU prediction branches in the training phase, they propose manually generating samples around ground truth.

In \cite{cai2018cascade, chen2019hybrid, qiao2020detectors}, they address the exponentially vanishing positive samples problem, utilizing three sequentially connected detectors to improve the hypothesis quality progressively, by resampling the ground truth.
It differs from our approach since we deal with the problem using a single detector and a segmentation head.
Authors of \cite{oksuz2020generating} give an interpretation about the fact that IoU imbalance negatively impacts the performance which is similar to ours. 
However, differently from us, they designed an algorithm to systematically generate the RoIs with required quality, where we base our work on the capabilities of the detector itself.

\noindent\textbf{Feature-level imbalance}.
A two-stage network deals with images containing objects of any size with the help of an FPN attached to the backbone.
How the RoI extraction layer combines the information provided by the FPN is of paramount importance to embody the highest amount of useful information.
This layer has been used by many derivative models such as Mask R-CNN, Grid \cite{lu2019grid}, Cascade R-CNN \cite{cai2019cascade}, HTC \cite{chen2019hybrid} and GC-net \cite{cao2019gcnet}.
In \cite{lin2017feature}, the authors apply an RoI pooling to a single and heuristically chosen FPN output layer.
However, as underlined by \cite{oksuz2020imbalance}, this method is defective due to a problem related to untapped information.
Authors of \cite{pont2016multiscale} propose to separately extract mask proposals from each scale and rescale them while including the results in a unique and multi-scale ranked list, selecting only the best ones. 
In \cite{ren2016object}, the authors use a backbone for each image scale, merging them with a max function.
On the contrary, we use an FPN which simplifies the network and avoids doubling the network parameters for each scale.

In SharpMask model \cite{pinheiro2016learning}, after making a coarse mask prediction, authors fuse feature layers back in a top-down fashion in order to reach the same size of the input image.
Authors of PANet \cite{liu2018path} point out that the information is not strictly connected with a single layer of the FPN. By propagating low-level features, they build another structure similar to FPN, coupled with it, combining the images pooled by the RoIs. 
While our proposed GRoIE layer is inspired by this approach, it differs from that in its size. We propose a novel way to aggregate data from the features pooled by RoIs making the network more lightweight without extra stacks coupled with FPN.

Auto-FPN \cite{xu2019auto} applies Neural Architecture Search (NAS) to the FPN.
PANet has been extended by AugFPN \cite{guo2020augfpn}.
The module with which we compare our module is called the Soft RoI Selector \cite{guo2020augfpn}, which includes an RoI pooling layer on each FPN layer to concatenate the results. Then, they are combined using the \emph{Adaptive Spatial Fusion} in order to build a weight map that is fed into 1x1 and 3x3 convolutions sequentially. In our work, we first carry out a distinct convolution operation on each output layer of the FPN network. After that, instead of concatenating, we sum the results since it is potentially more helpful for the network. In the end, we apply an attention layer whose job is to further filter the multi-scale context.

Authors of Multi-Scale Subnet \cite{linh2018multi} propose an alternative technique to RoI Align which employs cropped and resized branches for RoI extraction at different scales. In order to maintain the same number of outputs for each branch, they utilize convolutions with 1x1 kernel size, performing an average pooling to diminish them to the same size before summing them up. Finally, they use a convolutional layer with 3x3 kernel size as the post-processing stage. In our ablation study, we show that these convolutional configurations to carry out pre- and post-processing are not the optimal ones that can lead to better performance.

The IONet model \cite{bell2016inside} proposes doing away with any FPN network and, instead, using re-scaled, concatenated, and condensed (dimension-wise) features directly from the backbone before doing classification and regression. Finally, Hypercolumn \cite{hariharan2015hypercolumns} utilizes a hypercolumn representation to classify a pixel, with 1x1 convolutions and up-sampling the results to a common size so that they can be summed. Here, the absence of an optimized RoI pooling solution and an FPN layer and the simple processing of columns of pixels that have been taken from various stages of the backbone can be a limitation. In fact, we show in our ablation study that the adjacent pixels are necessary for optimal information extraction.

In \cite{tian2020conditional} they avoid to select the FPN layer and then RoI crop the features, attaching a convolutional branch on top of the last FPN layer and conditioning on the instance. In our case, we avoid the risk to loose information in intermediate FPN layers, leaving to the network the job of conditionally merging them for each instance.
\section{Self-Balanced R-CNN model}
\label{sec:sbrcnn}

In this section, we will describe our new architecture called Self-Balanced R-CNN (SBR-CNN), formed by three main contributions:
a $R^3$-CNN \cite{rossi2021recursively} enhanced version (subsection \ref{Recursively Refined R-CNN}),
the new FCC head architecture (subsection \ref{sec:FCC})
and the new GRoIE \cite{rossi2020novel} more performing version (subsection \ref{groie}).
Each of these contributions will be treated in detail individually.

\subsection{Recursively Refined R-CNN ($R^3$-CNN)}
\label{Recursively Refined R-CNN}

\begin{figure*}[bth]
	\begin{center}
		\includegraphics[trim=0 0 0 0, clip, width=0.6\linewidth]{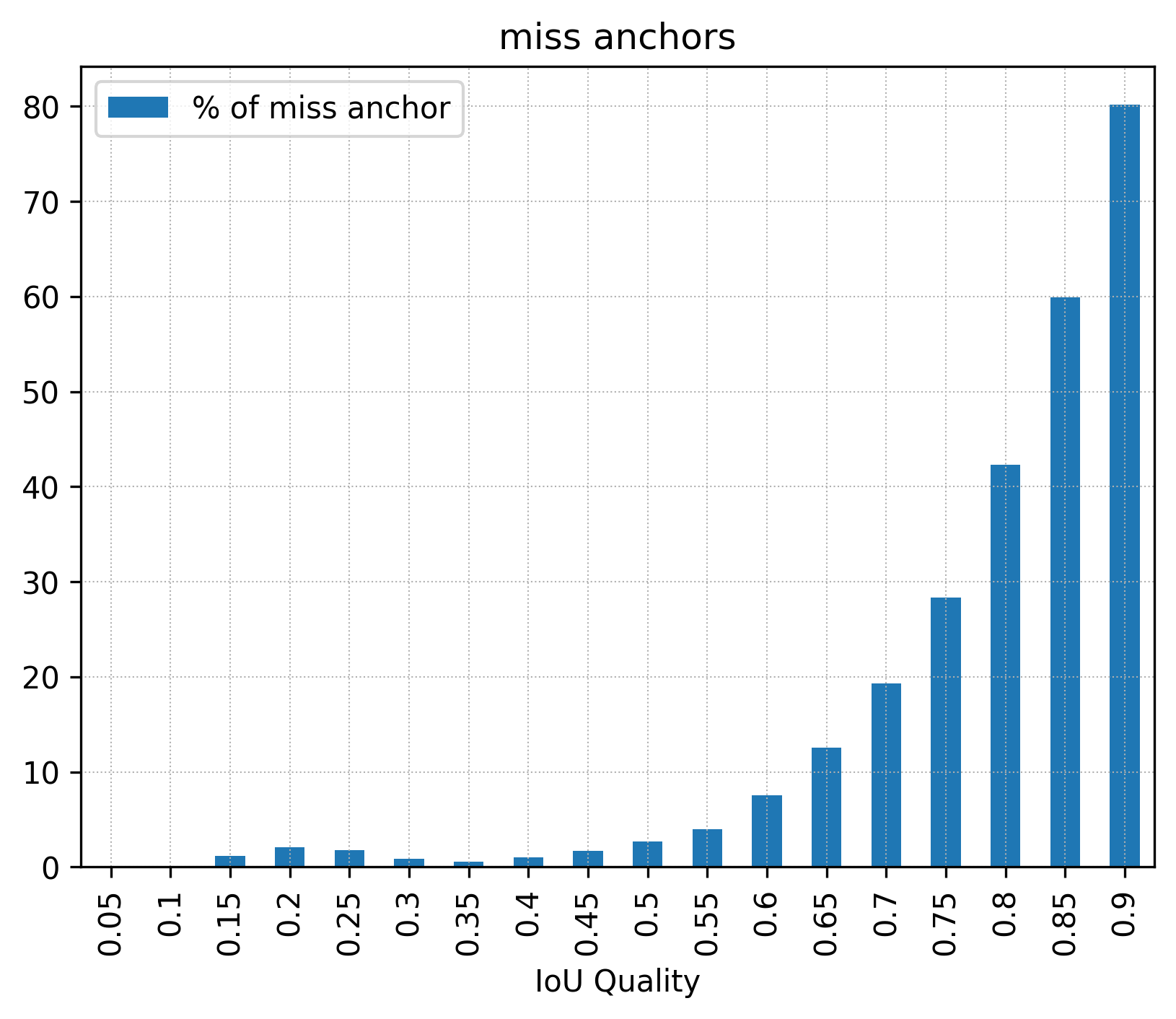}
	\end{center}
	\caption{Percentage of times in which, during the RPN training, there does not exist an anchor with a certain value of IoU w.r.t. the ground-truth bounding boxes.}
	\label{fig:miss-anchors}
\end{figure*}

In a typical two-stage network for instance segmentation, to obtain a good training of the network, we need as good candidates as possible from the RPN.
We could highlight at least two problems which are parts of so called IoU Distribution Imbalance (IDI) that afflict the training.
The first one, shown in Fig. \ref{fig:miss-anchors}, is related to the anchor system. It is called Exponential Vanishing Ground-Truth (EVGT) problem, where the higher the IoU threshold to label positive anchors is, the exponentially higher the percentage of missed ground-truth bounding boxes (gt-bboxes) can be.
For instance, more than 80\% of the gt-bboxes do not have a corresponding anchor with an IoU (w.r.t. the gt-bbox) between 0.85 and 0.9.
Since, for every image, the anchors' maximum IoU varies from one gt-bbox to another, if we choose a too high IoU threshold, some of the objects could be completely ignored during the training, reducing the number of truly used annotations.
For example, if the gt-bbox is in an unfortunate place where the maximum IoU between that and all available anchors is 0.55 and we choose a minimum threshold of 0.6, then no anchors will be associated with that object and it will be seen as part of the background during the training.
That is why we are usually obliged to use a very low threshold (typically 0.3 as a limit), since otherwise we could run into a case where a consistent part of the ground-truth is ignored.
The second, called Exponentially Vanishing Positive Samples (EVPS) problem \cite{cai2018cascade}, is partially connected with the first one because training the RPN with a too low threshold will reflect the low quality issue on its proposals.
Even in the best case, where each gt-bbox has the number of positive anchors greater than zero, the number of proposals from the RPN still diminishes exponentially with the increase of the required IoU threshold (see Fig. \ref{fig:mask}).

\begin{figure*}[bth]
	\begin{center}
		\includegraphics[trim=0 340 0 130, clip, width=1.1\linewidth]{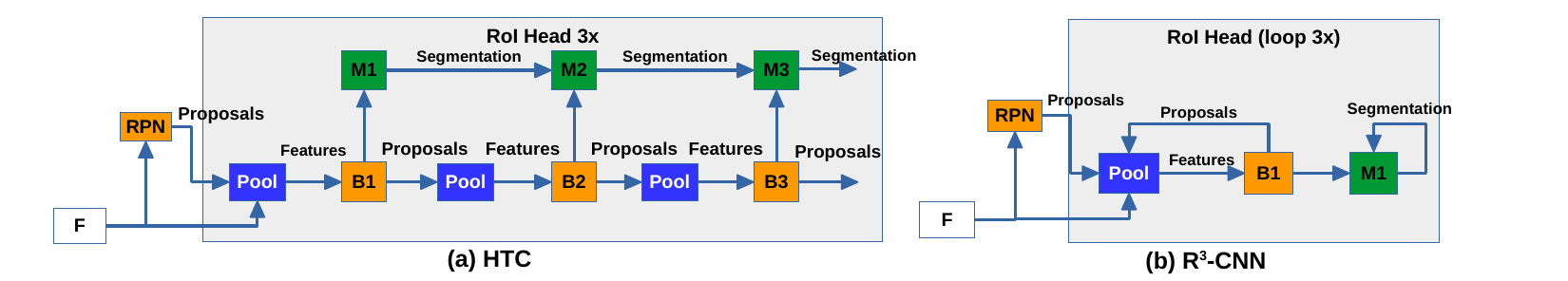}
	\end{center}
	\caption{Network design. (a) HTC: a multi-stage network which trains each head in a cascade fashion. (b) $R^3$-CNN: our architecture which introduces two loop mechanisms to self-train the heads.}
	\label{fig:recref-arch}
\end{figure*}

Fig. \ref{fig:recref-arch}(a) shows the Hybrid Task Cascade (HTC) model \cite{chen2019hybrid}, greatly inspired by Cascade R-CNN network \cite{cai2018cascade}, which trains multiple regressors connected sequentially, each of which is specialized in a predetermined and growing IoU minimum threshold.
This architecture offers a boost in performance at the cost of the duplicate heads, three times the ones used in Mask R-CNN.

In order to reduce the complexity, we designed a lighter architecture called Recursively Refined R-CNN ($R^3$-CNN) (see Fig. \ref{fig:recref-arch}(b)) to address the IDI problem by having single detection and mask heads trained uniformly on all the IoU levels.
In \cite{cai2018cascade}, it has been pointed out that the cost-sensitive learning problem \cite{elkan2001foundations, masnadi2010cost}, connected with the optimization of multiple IoU thresholds, needs multiple loss functions.
This encouraged us to look for a multiple selective training to address the problem.
The IoU threshold is used to distinguish between positive (an object) and negative (background) proposals.
Usually, because the Mask R-CNN-like architectures suffer from the EVGT and EVPS problems, the IoU threshold is set to 0.5, in order to have a good compromise between having enough samples to train the RoI head and to not degrading excessively the quality of the samples.
Because the maximum value for IoU is 1.0, we use a different and uniformly chosen IoU threshold in the range between 0.5 and 0.9 for each loop.
In this way, we sample a proposal list to feed the detector itself each time with a different IoU quality distribution.
We rely directly on RPN only in the initial loop.
Furthermore, this new list of proposals is used to feed the segmentation head M1, which incorporates an internal loop to refine the mask.

\begin{figure*}
\centering
\subfigure[Mask R-CNN trained 36 epochs]{\includegraphics[width=0.4\linewidth]{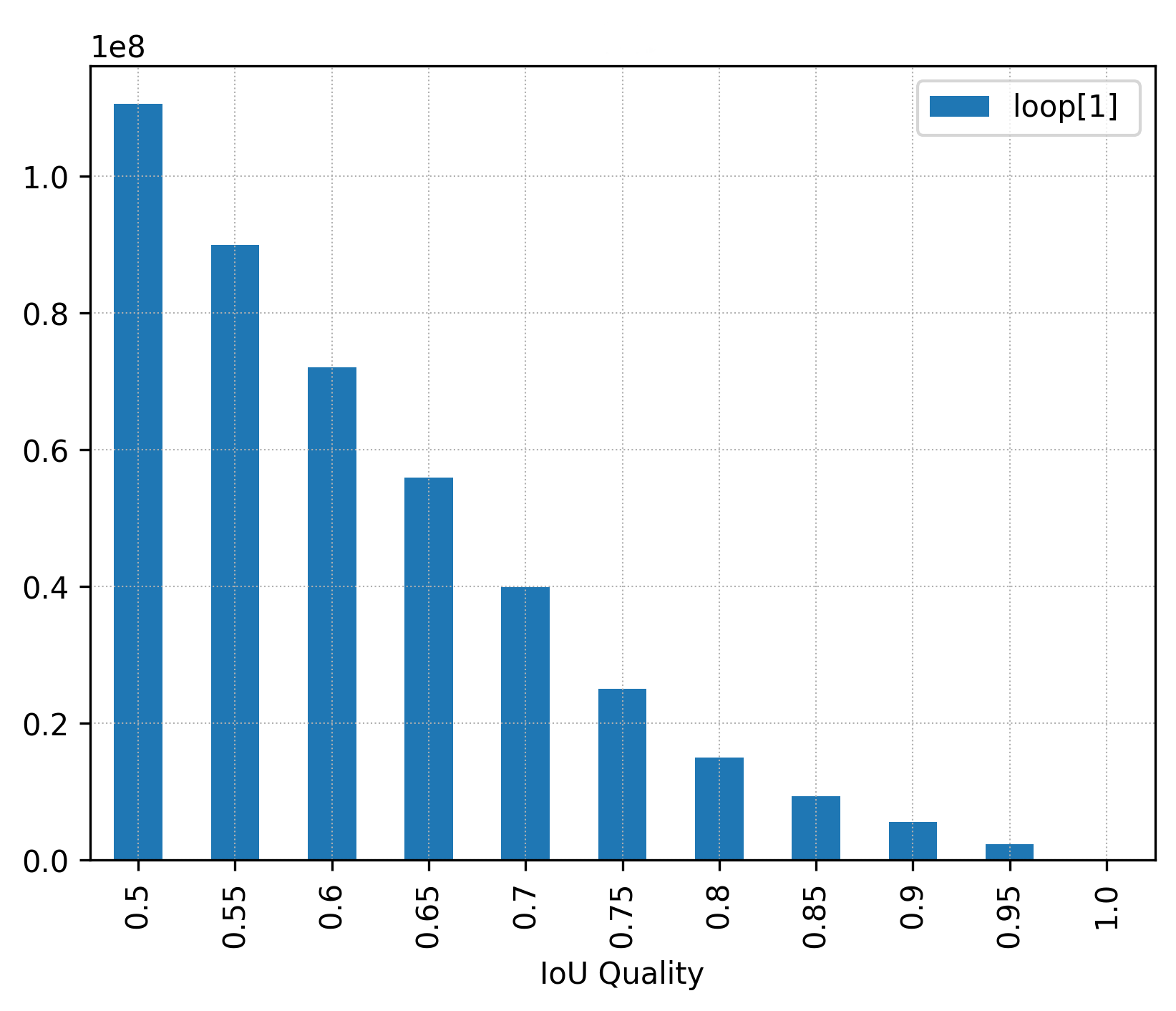}\label{fig:mask}}
\subfigure[$R^3$-CNN]{\includegraphics[width=0.4\linewidth]{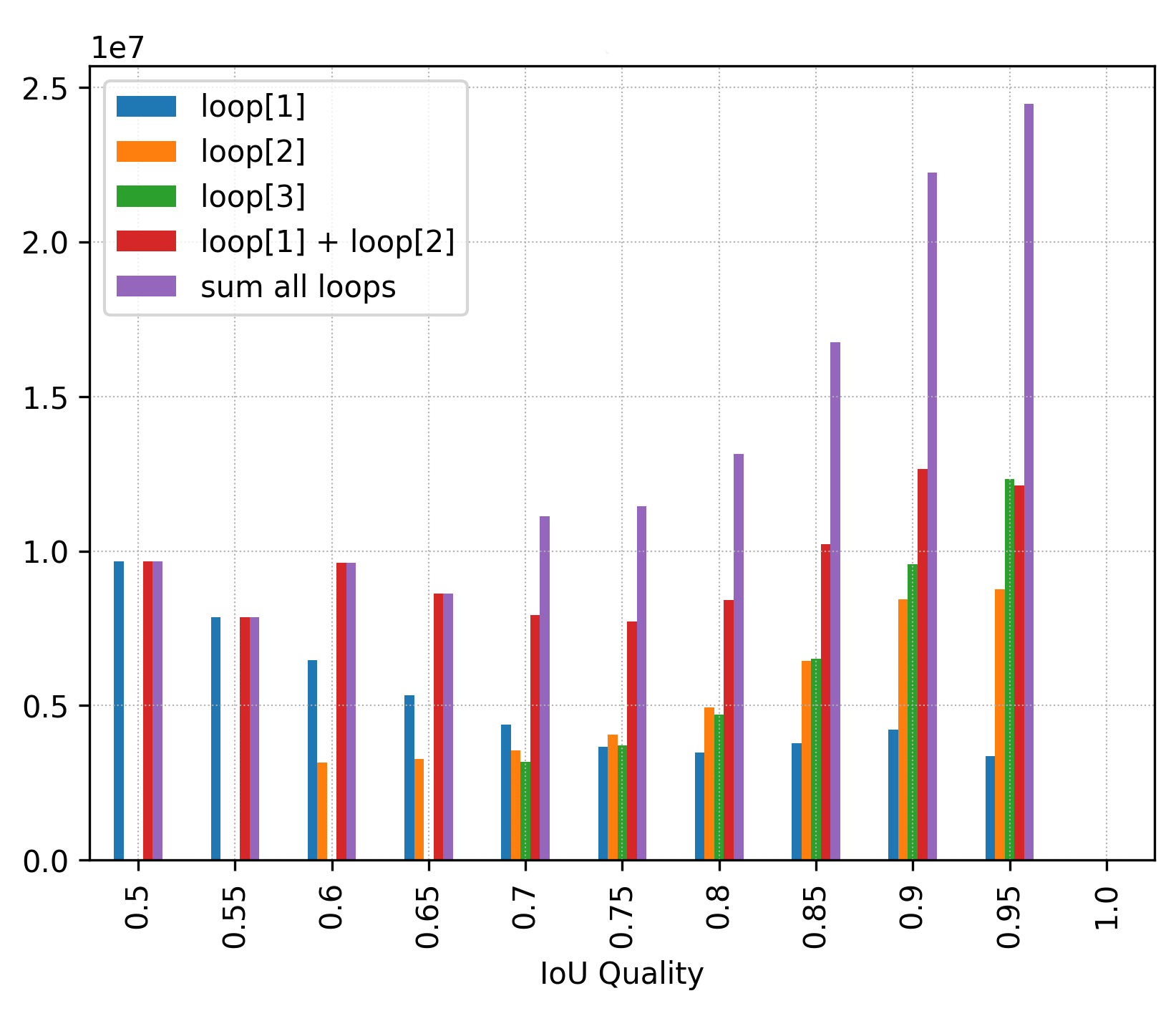}\label{fig:increase}}
	\caption{The IoU distribution of training samples for Mask R-CNN with a 3x schedule (36 epochs) (a), and $R^3$-CNN where at each loop it uses a different IoU threshold [0.5, 0.6, 0.7] (b). Better seen in color.}
	\label{fig:iou-histogram}
\end{figure*}

In order to show the rebalancing, during the training we collected the information of IoU of the proposals with the gt-bboxes.
To make the comparison fairer, we trained the Mask R-CNN three times the number of epochs.
In Fig. \ref{fig:mask}, we can see that the distribution of IoUs in Mask R-CNN maintains its exponentially decreasing trend.

Shown in Fig. \ref{fig:increase}, $R^3$-CNN presents a well-defined IoU distribution for each loop.
With two loops, we already have a more balanced trend and, by summing the third, the slope starts to invert.
The same trend can be observed in Cascade R-CNN, where the IoU histogram of the $n^{th}$ stage of Cascade R-CNN (as shown in \cite{cai2018cascade} - Fig. 4) can be compared with the $n^{th}$ loop of $R^3$-CNN.

Let us now define how the detection head loss (see Fig. \ref{fig:recref-arch}(b)) is composed, followed by the definition of the loss of the mask head.
For a given loop $t$, let us define B1 as the detection head, composed of $h$ as the classifier and $f$ as the regressor, which are trained for a selected IoU threshold $u^t$, with $u^t > u^{t-1}$.
Let $\mathbf{x^t}$ represent the extracted features from the input features $\mathbf{x}$ using the proposals $\mathbf{b^t}$.
In the first loop, the initial set of proposals ($\mathbf{b^0}$) comes from the RPN.
For the rest of the loops, in loop $t$, we have a set of $N_P$ proposals $\mathbf{b^t}=\left\{b_1^t, b_2^t,\dots,b_{N_P}^t\right\}$ obtained by the regressor $f$ using the extracted features $\mathbf{x^{t-1}}$ and the set of proposals $\mathbf{b^{t-1}}$ from the previous loop.

A given proposal $b_i^t \in \mathbf{b^t}$ is compared with all the $N_{GT}$ gt-bboxes $\mathbf{g}=\left\{g_1,g_2,\dots,g_{N_{GT}}\right\}$ by computing their overlap through the IoU. If none of these comparisons results in an IoU greater than the selected threshold $u^t$ for the current loop, the label $y_i^t=0$ corresponding to the class "background" is assigned to $b_i^t$. Otherwise, the label $l_x$ corresponding to the class of the gt-bbox $g_x$ with the maximum IoU is assigned to $y^t_i$:
\begin{equation}
l_x = \argmax{l_{\bar{x}}} IoU\left(b_i^t,g_{\bar{x}}\right)\:\:\:\:\:\forall g_{\bar{x}} \in \mathbf{g} | IoU\left(b_i^t,g_{\bar{x}}\right) \geqslant u^t
\end{equation}
\noindent where $l_{\bar{x}}$ is the label assigned to $g_{\bar{x}}$.
The detection head loss for the loop $t$ can be computed similarly as in Cascade R-CNN \cite{cai2018cascade}:

\begin{equation}
L_{bbox}^t\left(\mathbf{x^t}, \mathbf{g}\right) =
\begin{cases}
L_{cls}\left(h\left(\mathbf{x^t}\right), \mathbf{y^t}\right) & \forall b_i^t \:|\: y_i^t=0 \\
L_{cls}\left(h\left(\mathbf{x^t}\right), \mathbf{y^t}\right) + \lambda L_{loc}\left(f\left(\mathbf{x^t}, \mathbf{b^t}\right), \mathbf{g}\right) & \textrm{otherwise}
\end{cases}
\end{equation}

\noindent where $\mathbf{y^t}$ is the set of labels assigned to the proposals $\mathbf{b^t}$ and $\lambda$ is a positive coefficient.
The classification loss $L_{cls}$ is a multi-class cross entropy loss.
If $y_i^t$ is not zero, the localization loss $L_{loc}$ is also used, which is computed with a smooth L1 loss.

Regarding the segmentation branch performed by the M1 mask head, a separate RoI extraction module is employed to obtain the features $\mathbf{x^t}$ for the proposals $\mathbf{b^t}$ provided by the B1 detection head.
Similar to HTC, but with a single mask head, $R^3$-CNN uses an internal loop of $j$ iterations, with $j=t$, meaning that in the first loop of $R^3$-CNN, a single iteration ($j=1$) is performed, then two iterations in the second loop, and so forth.
At each internal iteration, the mask head receives as input the features $\mathbf{x^t}$ summed with the result of a $1\times1$ convolution $C_1$ applied to the output of the previous internal iteration:
\begin{equation}
\begin{split}
\mathbf{m^0} &= M1\left(\mathbf{x^t} + C_1(\mathbf{0})\right)\\
\mathbf{m^1} &= M1\left(\mathbf{x^t} + C_1(\mathbf{m^0})\right)\\
...&\\
\mathbf{m^{j-1}} &= M1\left(\mathbf{x^t} + C_1(\mathbf{m^{j-2}})\right)\\
\end{split}
\end{equation}

\noindent where $C_1$ is applied to a null tensor $\mathbf{0}$ at the first loop, and to the output of the previous iteration for the subsequent.
With this mechanism, the network iteratively refines its segmentation output.

The final output $\mathbf{m^{j-1}}$ of the internal loop is then upsampled with $U$ to reshape its size from $14\times14$ to $28\times28$. Finally, another $1\times1$ convolution $C_2$ is applied in order to reduce the number of channels to the number of classes:
\begin{equation}
\begin{split}
\mathbf{m^j} &= C_2\left(U\left(\mathbf{m^{j-1}}\right)\right)
\end{split}
\end{equation}
\noindent The loss function for the segmentation $L^t_{mask}$ is computed over $\mathbf{m^j}$ as follows:
\begin{equation}
	L_{mask}^t = BCE\left(\mathbf{m^j}, \mathbf{\hat{m}}\right)
\end{equation}

\noindent where $\mathbf{\hat{m}}$ represents the segmentation of the ground-truth object and $BCE$ is the binary cross entropy loss function.

In the end, the total loss for loop $t$ is composed as the sum of previous losses:
\begin{equation}
   L^t = \alpha_t \left(L_{bbox}^t + L_{mask}^t\right)
\end{equation}

\noindent where $\alpha_t$ represents a hyper-parameter defined statically in order to weight the different contributions of each loop.

We maintain the loop mechanism also at inference time and, at the end, we merge all the predictions, computing the average of the classification predictions.

\subsection{Fully Connected Channels (FCC)}
\label{sec:FCC}

In order to further reduce the network size, we propose to replace fully connected (FC) layers with convolutions.
In $R^3$-CNN model, they are included in two modules: in the detection head and in the Mask IoU branch \cite{huang2019mask}, which learns a quality score for each mask.

\begin{figure*}
	\begin{center}
		\includegraphics[width=0.6\linewidth]{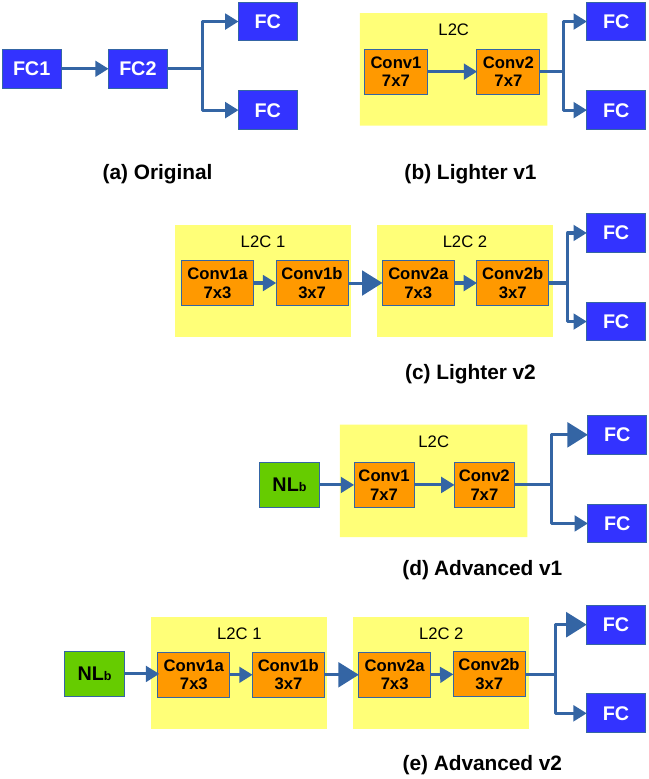}
	\end{center}
	\caption{(a) Original HTC detector head.
		     (b) Our lighter detector using convolutions with $7\times7$ kernels.
		     (c) Evolution of (b) with rectangular convolutions.
		     (d) Evolution of (b) with non-local pre-processing block.
		     (e) Evolution of (c) with non-local pre-processing block.
	     }
	\label{fig:l2c-arch}
\end{figure*}

In the detection head, the first two FC layers are shared between the localization and the classification tasks, followed by one smaller FC layer for each branch (see Fig. \ref{fig:l2c-arch}(a)).
Our goal is to replace the first two shared FC layers, which contain most of the weights, with convolutional layers, in order to obtain a lighter network (see Fig. \ref{fig:l2c-arch}(b)).
With the term $L2C$ we will refer, hereinafter, to these two convolutional layers together.
The input feature map has the shape of $n\times channels\times7\times7$ ($n$ is the number of proposals), characterized by a very small width and height.
A similar problem, addressed by \cite{rossi2020novel}, demonstrates how performance improves as the kernel size increases, covering almost the entire features shape.
We chose a large kernel size of $7 \times 7$ with padding 3 in order to maintain the input shape, halving the number of channels in input.
So, the first layer has 256 channels in input and 128 in output, while the second one reduces channels from 128 to 64.

Fig. \ref{fig:l2c-arch}(c) shows an alternative version which substitutes each convolution with two of them but with a small kernel ($7\times3$ with padding 3, and $3\times7$ with padding 1), with the aim of increasing the average precision and execution time.
Table \ref{tab:params-l2c} shows the sharp reduction obtained by the introduction of both $L2C$ versions.

\begin{table*}[t]
	\centering
	\begin{tabular}{|l|r|rl|}
		\hline
		Name  & \# Params & Description & \\
		\hline
		FC 1        & 12,846,080 & 256$\times$7$\times$7$\times$1024 (W) &+ 1024 (b) \\
		L2C (conv1) &  1,605,760 & 256$\times$7$\times$7$\times$128 (W)  &+ 128 (b) \\
		L2C (conv1a)&  1,376,512 & 256$\times$7$\times$3$\times$256 (W)  &+ 256 (b) \\
		L2C (conv1b)&    688,256 & 256$\times$3$\times$7$\times$128 (W)  &+ 128 (b) \\
		\hline
		FC 2        &  1,049,600 & 1024$\times$1024 (W)                  &+ 1024 (b) \\
		L2C (conv2) &    401,472 & 128$\times$7$\times$7$\times$64 (W)   &+ 64 (b)\\
		L2C (conv2a)&    344,192 & 128$\times$7$\times$3$\times$128 (W)  &+ 128 (b) \\
        L2C (conv2b)&    172,096 & 128$\times$3$\times$7$\times$64 (W)   &+ 64 (b) \\
		\hline
	\end{tabular}
	\caption{Parameter count for FC \& L2C with $7\times 7$ and rectangular kernels. W: weights; b: bias.}
	\label{tab:params-l2c}
\end{table*}

A heavier version of FCC includes also one non-local layer before the convolutions (see Fig. \ref{fig:l2c-arch}(d) and (e)).
Our non-local layer, differently from the original one \cite{wang2018non}, increases the kernels of internal convolutions from $1\times1$ to $7\times7$, in order to better exploit the information that is flowing inside the features in input.
The disadvantage of increased execution time could be alleviated in future versions, for instance, by using depth-wise convolutions \cite{sandler2018mobilenetv2} or similar mechanisms.

In terms of number of parameters, FCC architectures reduces them from 14M to 2.2M, 2.8M, 8.6M, and 9.2M if we use versions $b$, $c$, $d$, and $e$, respectively.

These changes in the architecture have been considered also for the Mask IoU module, which is composed of four convolutional layers followed by three FC layers.
Also in this case, the first two FC layers have been replaced, achieving the following weight reduction:
from 16.3M to 4.6M, 5.1M, 10.6M and 11.1M with version $b$, $c$, $d$ and $e$, respectively.

As previously noticed by \cite{wu2020rethinking}, the architecture is influenced by the task that it tries to solve.
In our case, we observed that convolutions can successfully substitute FC layers in all cases.
But, if the task involves a classification, a mechanism to preserve spatial sensitivity information is needed (with an enhanced non-local module).
Conversely, when the network learns a regression task, as for the Mask IoU branch, an attention module is not needed.
\subsection{Generic RoI Extraction Layer (GRoIE)}
\label{groie}

The FPN is a commonly used architecture for extracting features from different image resolutions without separately elaborating each scale.
In a two-stage detection framework similar to one mentioned in this paper, the output layer of an FPN network is chosen heuristically as a unique source of sequential RoI pooling.
However, while the formula has been designed very well, it is obvious that the layer is selected arbitrarily.
Furthermore, the mere combination of the layers that are provided by the backbone can result in a non-uniform distribution of low- and high-level information in the FPN layers \cite{oksuz2020imbalance}.
This phenomenon necessitates finding a way to avoid losing information by selecting only one of them as well as
correctly combining them in order to obtain a re-balanced distribution.
The enhancement obtained from the GRoIE \cite{rossi2020novel} suggests that if all the layers are aggregated appropriately through some extra convolutional filters, it is more likely to produce higher quality features.
The goal is to solve the feature imbalance problem of FPN by considering all the layers, leaving the task of learning the best way of aggregating them to the network.

\begin{figure*}[t]
	\begin{center}
		\includegraphics[width=0.8\linewidth]{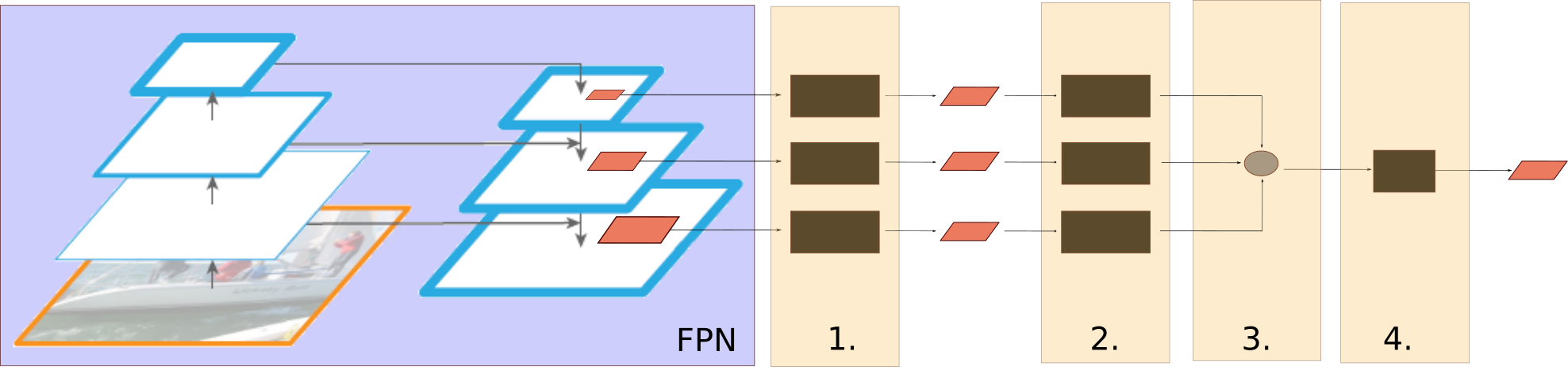}
	\end{center}
	\caption{GRoIE framework. (1) RoI Pooler. (2) Pre-processing phase. (3) Aggregation function. (4) Post-processing phase.}
	\label{fig:groi}
\end{figure*}

The original RoI Extraction Layer architecture is composed only by a RoI Pooler and a mathematical function to select the FPN layer on which apply the RoI Pooler to extract the features.
In Figure \ref{fig:groi}, the GRoIE four-stage architecture is shown.
Given a proposed region, a fixed-size RoI is pooled from each FPN layer (stage 1).
Then, the $n$ resulting feature maps, one for each FPN layer, are pre-processed separately (stage 2) and summed together (stage 3) to form a single feature map.
In the end, after a post-processing stage (stage 4), global information is extracted.
The pre- and post-processing stages are composed of a single or multiple layers, depending on the configuration which provides the best performance (see experimental section for details).
These could be formed by a simple convolutional layer or a more advanced attention layer like Non-local block \cite{wang2018non}.

The GRoIE architecture guarantees an equal contribution of all scales, benefiting from the embodied information in all FPN layers and overcoming the limitations of choosing an arbitrary FPN layer.
This procedure can be applied to both object detection and instance segmentation.
Our work focused on even improving the GRoIE model and evaluating new building blocks for the pre- and the post-processing stages.
In particular, as we did for the FCC, we tested bigger and rectangular kernels for the convolutional layers, to better exploit the close correlation between neighboring features.
The advantage to involve near features is even more evident when applied to a more sophisticated non-local module, which includes an attention mechanism.
However, as we will see in the ablation study, it is extremely important to do it in the right point of the chain.

\section{Experiments}
\label{experiments}

This section reports the extensive experiments carried out to demonstrate the effectiveness of the proposed architecture. After introducing the dataset, the evaluation metrics, the implementation details and the table legend, the following subsections report the results on the three main novelties of the architecture, namely the Recursively Refined R-CNN ($R^3$-CNN), the Fully Connected Channels (FCC) and the Generic RoI Extraction layer (GRoIE). Finally, the last subsection shows how all these novelties together can bring performance benefits to several state-of-the-art instance segmentation architectures.

\subsection{Dataset and Evaluation Metrics}
\label{sec:dataset-metrics}

\noindent\textbf{Dataset}.
As the majority of recent literature on instance segmentation, we perform our tests on the MS COCO 2017 dataset \cite{lin2014microsoft}.
The training dataset consists of more than 117,000 images and 80 different classes of objects.

\noindent\textbf{Evaluation Metrics}.
We used the same evaluation functions offered by the python \textit{pycocotools} software package, performed on the COCO minival 2017 validation dataset, which contains 5000 images.

We report the mean Average Precision (AP) for both bounding box ($B_{AP}$) and segmentation ($S_{AP}$) tasks.
The primary metric $AP$ is computed as average over results with IoU thresholds from 0.5 to 0.95.
Other metrics include $AP_{50}$ and $AP_{75}$ with 0.5 and 0.75 minimum IoU thresholds, respectively.
Separate metrics are calculated for small ($AP_s$), medium ($AP_m$) and large ($AP_l$) objects.

\subsection{Implementation Details}
\label{sec:implementation-details}

In order to perform a fair comparison, we use same the hardware and software configuration to carry out the experiments.
When available, the original code released by the authors was used.
Otherwise, we used the corresponding implementations in MMDetection \cite{mmdetection} framework.
In the case of HTC, we do not consider the semantic segmentation branch.

Unless mentioned otherwise, the following configuration has been used.
We performed a distributed training on 2 servers, each one equipped with 2x16 IBM POWER9 cores and 256 GB of memory plus 4 x NVIDIA Volta V100 GPUs with Nvlink 2.0 and 16GB of memory.
Each training consists of 12 epochs with Stochastic Gradient Descent (SGD) optimization algorithm, batch size 2 for each GPU, an initial learning rate of 0.02, a weight decay of 0.0001, and a momentum of 0.9. The steps to decay the learning rate was set at epochs 8 and 11.
Regarding the images, we fixed the long edge and short edge of the images to 1333 and 800, maintaining the aspect ratio.
ResNet50 \cite{he2016deep} was used as the backbone.

\subsection{Table Legend}
\label{sec:table-legends}

To ease the understanding of the following tables, we shortly introduce the notation used.
Since $R^3$-CNN has the loop both in training and evaluation phase, we denote the number of training and evaluation loops as $L_t$ and $L_e$, respectively. Whenever only $L_t$ is reported, $L_e$ is intended to have the same value of $L_t$.
In the case of HTC, $L_t$ corresponds to the number of stages.

The column $H$ (heads) specify how many pairs of detection (B) and mask (M) heads are included.
In the case of multiple pairs ($H > 1$), the column Alt. (alternation) gives information about which one is used for each loop.
For example, in row \#3 of Table \ref{tab:compare-alternacy-two-heads}, the model is using three loops for training and evaluation, and two pairs of B and M.
The column \textit{Alt} reports "abb", meaning that B1 and M1 are used only in the first loop, while B2 and M2 are used for the second and third loops.

The columns $M_{IoU}$, $L2C$, $NL_b$, $NL_a$ are flags indicating the presence of the Mask IoU branch with the associated loss, the substitution of the FC layers with convolutions ($L2C$), inside the detection head (in Table \ref{tab:compare-3-stages-1-head-Shared2NLFCBBoxHead}) or Mask IoU branch (in Table \ref{tab:compare-MaskIoU-modded}), and finally, the introduction of our non-local blocks with kernels $7\times7$ before ($NL_b$) and after ($NL_a$) the $L2C$ convolutions.
The column Speed refers to the number of processed images per second on evaluation phase with batch size equal to one and one GPU.


\subsection{Results for Recursively Refined R-CNN ($R^3$-CNN)}
\label{experiments Recursively Refined R-CNN}

\subsubsection{Preliminary analysis of $R^3$-CNN}


\noindent\textbf{Description}.
We compared Mask R-CNN and HTC with two $R^3$-CNN models: naive (one pair of bounding box B and mask M) and deeper (two pairs, with the alternation \textit{aab}).
To carry out a fair comparison, the Mask R-CNN was trained with 36 epochs instead of 12, and the optimal configuration for the HTC network was used.

\begin{table*}[t]
	\setlength{\tabcolsep}{3.5pt}
	\begin{center}
		\begin{tabular}{c|l|c|c|c|c|c|c|c|c}
			\# & Model & \# Params & $L_t$ & $H$ & $B_{AP}$ & $S_{AP}$ & Mem & Model & Speed \\
			\hline\hline
			1 & Mask (1x)             & 44,170 K & 1 & 1 & 38.2 & 34.7 & 4.4G & 339M & 11.6 \\
			2 & Mask (3x)             & 44,170 K & 1 & 1 & 39.2 & 35.5 & 4.4G & 339M & 11.6 \\
			3 & HTC                   & 77,230 K & 3 & 3 & 41.7 & 36.9 & 6.8G & 591M &  3.3 \\
			\hline
			4 & $R^3$-CNN (naive)     & 43,912 K & 3 & 1 & 40.9 & 37.2 & 6.7G & 337M &  3.4 \\
			5 & $R^3$-CNN (deeper)    & 60,604 K & 3 & 2 & 41.8 & 37.5 & 7.0G & 464M &  3.4 \\
		\end{tabular}
	\end{center}
	\caption{Comparison between $R^3$-CNN, Mask R-CNN, and HTC. Column \textit{Model} contains the number of parameters (millions). $3x$ means training with 36 epochs.}
	\label{tab:count_parameters}
\end{table*}

\noindent\textbf{Results}.
The \textit{naive} version has the biggest reduction in terms of the number of parameters, loosing $0.8$ in $B_{AP}$ but gaining $0.3$ in $S_{AP}$ compared to HTC.
Regarding the \textit{deeper} version, it matches the HTC in $B_{AP}$ and further increases the gap in $S_{AP}$, while still saving a considerable number of parameters.
Both of them require the same amount of memory, as well as the inference time as HTC.
This is due to the fact that the training procedure and the utilized components are very similar to those of HTC.

Compared to Mask R-CNN, our $R^3$-CNN (in both versions) outperforms it, even when Mask R-CNN is trained for a triple number of epochs (row \#2).
This can be explained by the very different way of training the network, helping to achieve a higher quality of the bounding boxes during the training. 
Moreover, the training phase for $R^3$-CNN is faster than Mask R-CNN (about 25 hours versus 35 hours), although $R^3$-CNN has the disadvantage of requiring the loop mechanism also in the evaluation phase.
\subsubsection{Ablation study on the training phase}


\begin{table*}[b]
	\begin{center}
		\begin{tabular}{c|l|c|c|c|c}
			\# & Model & $L_t$ & $H$ & $B_{AP}$ & $S_{AP}$ \\
			\hline\hline
			1 & Mask & 1 & 1 & 38.2 & 34.7  \\
			\hline
			2 & \multirow{4}{*}{$R^3$-CNN}     & 1 & 1 & 37.6 & 34.6 \\
			3 &           & 2 & 1 & 40.4 & 36.7 \\
			\one{4} &     & \one{3} & \one{1} & \one{40.9} & \one{37.2} \\
			5 &           & 4 & 1 & 40.9 & 37.4 \\
		\end{tabular}
	\end{center}
	\caption{Impact of the number of training loops in a $R^3$-CNN. Row \#4 is the naive $R^3$-CNN in Table \ref{tab:count_parameters}.}
	\label{tab:compare-num-training-stages}
\end{table*}

\noindent\textbf{Description}.
In these experiments, the network is trained with a number of loops varying from 1 to 4.
The number of loops for the evaluation changes accordingly. 
The basic architecture for all the tests in these experiments is the naive $R^3$-CNN with single pair of detection and mask heads.
It means that all the $R^3$-CNN models have the same number of parameters but they are trained more if the number of loop increases.\\

\noindent\textbf{Results}.
The results are reported in Table \ref{tab:compare-num-training-stages}.
Using a single loop (row \#2) not only produces a similar IoU distribution to Mask R-CNN as mentioned in Section \ref{Recursively Refined R-CNN}, but also leads to a similar performance.
With two loops (row \#3), we can reach almost the peak performance of $R^3$-CNN thanks to the rebalancing of IoU, surpassing the performance of Mask R-CNN.
In the case of three loops, the network provides more high-quality proposals, reaching even better performance on both tasks.
Adding four loops for training does not improve object detection task but still improves segmentation.
\subsubsection{Ablation study on the evaluation phase}


\begin{table*}[b]
	\begin{center}
		\begin{tabular}{c|l|c|c|c|c|c}
			\# & Model & $L_t$ & $H$ & $L_e$ & $B_{AP}$ & $S_{AP}$ \\
			\hline\hline
			1 & Mask       & 1 & 1 & 1 & 38.2 & 34.7  \\
			\hline
			2 & \multirow{5}{*}{$R^3$-CNN}     & 3 & 1 & 1 & 37.7 & 35.1  \\
			3 &          & 3 & 1 & 2 & 40.5 & 36.9  \\
			\one{4} &    & \one{3} & \one{1} & \one{3} & \one{40.9} & \one{37.2} \\
			5 &          & 3 & 1 & 4 & 40.8 & 37.2 \\
			6 &          & 3 & 1 & 5 & 40.9 & 37.3 \\
		\end{tabular}
	\end{center}
	\caption{Impact of evaluation loops $L_e$ in a three-loop and one-head-per-type $R^3$-CNN model. Row \#4 is the naive $R^3$-CNN in Table \ref{tab:count_parameters}.}
	\label{tab:compare-num-eval-stages}
\end{table*}

\noindent\textbf{Description}.
In this experiment we focus on how the results are affected by the number of loops in the evaluation phase.
We consider the \textit{naive} architecture mentioned above as the pre-trained model and we vary the number of evaluation loops.\\

\noindent\textbf{Results}.
From Table \ref{tab:compare-num-eval-stages}, we observe that we can not avoid to use the loop in the evaluation phase, because it plays the role to provide high quality RoIs to the network.
Though, already with two loops the AP values are significantly better (row \#3).
From four loops onward, the performance tends to remain almost stable in both detection and segmentation tasks.
\subsubsection{Ablation study on a two-heads-per-type model}


\begin{table*}[bth]
	\begin{center}
		\begin{tabular}{c|c|c|c|l|c|r}
			\# & Model & $L_t$ & $H$ & Alt. & $B_{AP}$ & $S_{AP}$ \\
			\hline\hline
			1 & HTC          & 3 & 3 & abc      & 41.7 & 36.9  \\
			\hline
			2 & \multirow{8}{*}{$R^3$-CNN}    & 2 & 2 & ab     & 40.9 & 36.5  \\
			3 &           & 3 & 2 & abb    & 41.8 & 37.2  \\
			4 &           & 3 & 2 & aab    & 41.8 & 37.5 \\
			5 &           & 3 & 2 & aba    & 41.5 & 37.2 \\
			\one{6} &     & \one{4} & \one{2} & \one{aabb}   & \one{42.1} & \one{37.7} \\
			7 &           & 4 & 2 & abab   & 41.9 & 37.6 \\
			8 &           & 5 & 2 & aabbb  & 41.8 & 37.5 \\
		\end{tabular}
	\end{center}
	\caption{The impact of the number of training loops and pair alternation in two-heads-per-type (two pairs B/M) in the $R^3$-CNN.}
	\label{tab:compare-alternacy-two-heads}
\end{table*}

\noindent\textbf{Description}.
In this experiment, we evaluate the performance on changing the number of loops and the alternation between the pairs of heads in the architecture.
It is worth emphasizing that increasing the number of loops does not change the number of weights.\\

\noindent\textbf{Results}.
Table \ref{tab:compare-alternacy-two-heads} reports the results.
In case of two loops (row \#2), the model shows good precision, but still worse than HTC.
With three loops and \textit{aab} alternation (row \#4), $R^3$-CNN surpasses HTC in both task.

With four loops (rows \#6 and \#7), the performances are all higher than HTC, especially for \emph{aabb} alternation (row \#6).
Finally, with five (row \#8) loops the performance is not increasing anymore.


\subsection{Results for Fully Connected Channels (FCC)}
\label{experiments fcc}

\subsubsection{Ablation study on the Detection Head}
\label{sec:ablation-detection-head}

\noindent\textbf{Description}. 
In this section, we evaluate the effect of the head redesign toward a fully convolutional approach.
We tested both $L2C$ versions (see Fig. \ref{fig:l2c-arch} (b-d) and Fig. \ref{fig:l2c-arch}(c-e) in orange) and the introduction of the non-local layer with larger kernels before (column $NL_b$) the $L2C$ convolutions (see Figure \ref{fig:l2c-arch}(d) and (e)) and, to have a more complete ablation study, also after them (column $NL_a$).
In order to provide a more comprehensive analysis, the case of two heads per type (column $H$) and four loops during training (column $L_t$) were also considered.

\noindent\textbf{Results}.
Table \ref{tab:compare-3-stages-1-head-Shared2NLFCBBoxHead} summarizes the results. 
As expected, the presence of only $L2C$ (see Fig. \ref{fig:l2c-arch}(b)) has an impact on performance (see row \#2 vs \#3).
Rectangular convolutions (row \#4 and Fig. \ref{fig:l2c-arch}(c) and (e)) help to almost completely mitigate this loss, approaching the original performance (row \#2), but with the advantage of lowering the number of parameters and speeding up the execution compared to row \#3.

\begin{table*}[bth]
	\begin{center}
		\begin{tabular}{c|l|c|c|c|c|c|c|c|c|c}
			\# & Model & $L_t$  & $H$ & $L2C$ & $NL_b$ & $NL_a$ & $B_{AP}$ & $S_{AP}$ & Speed & \# Params (M) \\
			\hline\hline
			1 & HTC                        & 3 & 3 &                                    &            &            & 41.7 & 36.9 & 3.3 & 77.2 \\
			\hline
			2 & \multirow{6}{*}{$R^3$-CNN} & 3 & 1 &                                    &            &            & 40.9 & 37.2 & 3.4 & 43.9 \\
			3 &                            & 3 & 1 & $7 \times 7$                       &            &            & 39.8 & 36.4 & 2.2 & 32.2 \\
			4 &                            & 3 & 1 & $7\times 3 \rightarrow 3 \times 7$ &            &            & 40.6 & 36.8 & 2.8 & 32.7 \\
			5 &                            & 3 & 1 & $7 \times 7$                       & \checkmark &            & 41.8 & 37.6 & 1.0 & 38.6 \\
			\one{6} &                      & \one{3} & \one{1} & \one{$7\times 3 \rightarrow 3 \times 7$} & \one{\checkmark} &  & \one{41.7} & \one{37.6} & \one{1.2} & \one{37.9} \\
			7&                             & 3 & 1 & $7 \times 7$                       & \checkmark & \checkmark & 41.8 & 37.6 & 0.9 & 39.0 \\
			\hline
			8& \multirow{5}{*}{$R^3$-CNN}  & 4 & 2 &                                    &            &            & 41.9 & 37.5 & 2.9 & 60.6 \\
			9&                             & 4 & 2 & $7 \times 7$                       &            &            & 41.4 & 37.2 & 1.8 & 37.1 \\
			10&                            & 4 & 2 & $7\times 3 \rightarrow 3 \times 7$ &            &            & 41.0 & 37.1 & 2.4 & 38.3 \\
			\one{11} &                     & \one{4} & \one{2} & \one{$7 \times 7$}     & \one{\checkmark} &      & \one{42.9} & \one{38.1} & \one{0.8} & \one{50.0} \\
			12&                            & 4 & 2 & $7\times 3 \rightarrow 3 \times 7$ & \checkmark &            & 42.6 & 37.8 & 0.9 & 51.1 \\
		\end{tabular}
	\end{center}
	\caption{Impact of FCC module configurations applied to $R^3$-CNN detector. Row \#2 is the $R^3$-CNN in row \#4 of Table \ref{tab:count_parameters}.}
	\label{tab:compare-3-stages-1-head-Shared2NLFCBBoxHead}
\end{table*}

The non-local block before $L2C$ (row \#5) boosts the performance, matching $B_{AP}$ of HTC and surpassing its $S_{AP}$ by a good margin. 
Conversely, its introduction after $L2C$ does not bring any benefits.

In the case of two heads per type and four loops, $L2C$ produces higher performance (see rows \#9 and \#8) compared to row \#3.
Rectangular convolutions (row \#10) worsen the performance compared to row \#9, but have the advantage of a good increase in speed.
As in the previous case, the introduction of our non-local module (row \#11 and \#12) produces a good performance boost with respect to the model without them (row \#9 and \#10).

To summarize, FCC with only $L2C$ makes the network lighter, reducing the wight by 14 to 18 percent, while slightly worsening the performance compared to using FC layers.
Moreover, a boost in performance is achieved by the non-local block inserted before $L2C$, surpassing the original performance with a good margin, albeit at the cost of a higher execution time.

\subsubsection{Ablation study on Mask IoU module}
\label{sec:maskiou}

\noindent\textbf{Description}.
In order to increase performance even further, we borrowed the Mask IoU learning task from \cite{huang2019mask} and redesigned its branch to introduce as few weights as possible.
After testing the original Mask IoU branch, as done previously on detection head, we conducted an ablation study.
We considered two baselines: a lighter (row \#2) and a better-performing (row \#8) model in Table \ref{tab:compare-MaskIoU-modded}.
They also correspond to rows \#6 and \#11 in Table \ref{tab:compare-3-stages-1-head-Shared2NLFCBBoxHead}, respectively.\\

\begin{table*}[bth]
	\begin{center}
		\begin{tabular}{l|l|c|c|c|c|c|c|c|c|c|c}
			\# & Model      & $L_t$ & $H$ & $M_{IoU}$ & $L2C$ & $NL_b$ & $NL_a$ & $B_{AP}$ & $S_{AP}$ & Speed & \# Params (M) \\
			\hline\hline
			1 & HTC                        & 3 & 3 &            &                                     &            &             & 41.7 & 36.9 & 3.3 & 77.2 \\
			\hline
			2 & \multirow{6}{*}{$R^3$-CNN} & 3 & 1 &            &                                     &            &             & 41.7 & 37.6 & 1.2 & 37.9 \\
			3 &                            & 3 & 1 & \checkmark &                                     &            &             & 41.6 & 38.5 & 1.1 & 54.9 \\
			\one{4} &                      & \one{3} & \one{1} & \one{\checkmark} & \one{$7 \times 7$}&            &             & \one{41.7} & \one{38.4} & \one{1.1} & \one{43.2} \\
			5 &                            & 3 & 1 & \checkmark & $7\times 3 \rightarrow 3 \times 7$  &            &             & 41.8 & 38.4 & 1.1 & 44.3 \\
			6&                             & 3 & 1 & \checkmark & $7 \times 7$                        & \checkmark &             & 41.4 & 38.3 & 1.1 & 49.6 \\
			7&                             & 3 & 1 & \checkmark & $7 \times 7$                        & \checkmark & \checkmark  & 41.6 & 38.3 & 1.1 & 50.0 \\
			\hline
			8& \multirow{5}{*}{$R^3$-CNN}  & 4 & 2 &            &                                     &             &            & 42.9 & 38.1 & 0.8 & 50.0 \\
			9&                             & 4 & 2 & \checkmark &                                     &            &             & 42.7 & 38.6 & 0.9 & 66.3 \\
			\one{10}&                  & \one{4} & \one{2} & \one{\checkmark} & $7 \times 7$ & & & \one{42.7} & \one{38.7} & \one{0.8} & \one{54.6} \\
			11&                      & 4 & 2 & \checkmark & $7 \times 7$& \checkmark &       & 42.8 & 38.7 & 0.8 & 61.0 \\
			12&                            & 4 & 2 & \checkmark & $7 \times 7$                        & \checkmark & \checkmark  & 42.7 & 38.6 & 0.8 & 61.4 \\
		\end{tabular}
	\end{center}
	\caption{Impact of FCC to Mask IoU branch.}
	\label{tab:compare-MaskIoU-modded}
\end{table*}

\noindent\textbf{Results}.
Table \ref{tab:compare-MaskIoU-modded} summarizes the results.
As expected, original Mask IoU module (rows \#3 and \#9) improves the segmentation.
Differently from the detection head, the redesigned Mask IoU branch with only $L2C$ with $7\times7$ kernels (rows \#4 and \#10) is enough to maintain almost the same performance compared to the original branch (rows \#3 and \#9), but introduces few new parameters and almost does not affect the execution time.
Contrary to the previous experiment, neither rectangular convolutions (row \#5) nor our non-local blocks (rows \#6 and \#7) bring any noticeable improvement.
\subsection{Results on Generic RoI Extractor (GRoIE)}
\label{groie_component ablation study}

For the following experiments, we chose the Faster R-CNN as the \textit{baseline} to have a generic and lightweight model to compare with.
Our goal, for the following experiments, is to find the best layers for the pre- and post-processing.
Conv $3 \times 3$, $5 \times 5$, $7 \times 7$ mean we are using 2D convolution with kernel $3 \times 3$, $5 \times 5$, $7 \times 7$ , respectively.
Conv $7 \times 3 \rightarrow 3 \times 7$ means that we use two consecutive 2D convolutional layers with $7 \times 3$ and $3 \times 7$, respectively.
For the Non-local block \cite{wang2018non}, we tested the original architecture composed by convolutional layers with kernel $1 \times 1$ and a customized version, composed by convolutional layers with kernel $7 \times 7$.

\subsubsection{Pre-processing module analysis}

\noindent\textbf{Description}.
For this ablation analysis, we did not apply any post-processing.
We tested two types of pre-processing: a convolutional layer with different kernel sizes and a non-local block.\\

\begin{table*}[bth]
	\begin{center}
		\begin{tabular}{l|c|c|c|c|c|c}
			Method & $AP$ & $AP_{50}$ & $AP_{75}$ & $AP_{s}$ & $AP_{m}$ & $AP_{l}$ \\
			\hline\hline
			baseline                  & 37.4 & 58.1 & 40.4 & 21.2 & 41.0 & 48.1 \\
			Conv $3 \times 3$         & 38.1 & 58.7 & 41.5 & 22.2 & 41.7 & 49.0 \\
			Conv $5 \times 5$         & 38.2 & 59.2 & 41.6 & 22.5 & 41.6 & 49.0 \\
			\one{Conv $7 \times 7$}         & \one{38.3} & \one{59.2} & \one{41.6} & \one{22.7} & \one{41.7} & \one{49.4} \\
			Conv $7 \times 3 \rightarrow 3 \times 7$ & 37.9 & 58.5 & 41.3 & 22.0 & 41.5 & 49.1 \\
			Non-local $1 \times 1$    & 37.7 & 58.9 & 40.7 & 22.0 & 41.4 & 48.5 \\
			Non-local $7 \times 7$    & 38.4 & 59.2 & 41.9 & 22.5 & 42.1 & 49.5 \\
		\end{tabular}
	\end{center}
	\caption{Ablation analysis on pre-processing module.}
	\label{ablation-preprocessing}
\end{table*}

\noindent\textbf{Results}.
Table \ref{ablation-preprocessing} shows the results.
The increase in the kernel size improves the final performance, confirming the close correlation between neighboring features.
The use of a rectangular convolution did not help as it did in Section \ref{sec:ablation-detection-head} for the detection head.
In the case of the non-local module, the original one does not have the expected benefit.
Our non-local module with a larger kernels gives a slight advantage over the others, but not enough to justify the introduced slowdown.

\subsubsection{Post-processing module analysis}

\noindent\textbf{Description}.
In this experiment we analyze the post-processing module, by not applying any pre-processing.\\

\begin{table*}[bth]
	\begin{center}
		\begin{tabular}{l|c|c|c|c|c|c}
			Method & $AP$ & $AP_{50}$ & $AP_{75}$ & $AP_{s}$ & $AP_{m}$ & $AP_{l}$ \\
			\hline\hline
			baseline                    & 37.4 & 58.1 & 40.4 & 21.2 & 41.0 & 48.1 \\
			Conv $3 \times 3$           & 37.3 & 58.3 & 40.4 & 21.2 & 41.0 & 48.5 \\
			Conv $5 \times 5$           & 37.8 & 58.7 & 40.9 & 22.2 & 41.2 & 48.8 \\
			Conv $7 \times 7$           & 37.9 & 59.0 & 41.2 & 21.5 & 41.8 & 48.6 \\
			Conv $7 \times 3 \rightarrow 3 \times 7$ & 37.4 & 58.4 & 40.5 & 21.4 & 40.9 & 48.7 \\
			Non-local $1 \times 1$       & 37.8 & 59.1 & 40.5 & 22.0 & 41.7 & 48.3 \\
			\one{Non-local $7 \times 7$} & \one{38.7} & \one{59.7} & \one{42.3} & \one{22.7} & \one{42.4} & \one{49.7} \\
		\end{tabular}
	\end{center}
	\caption{Ablation analysis on post-processing module.}
	\label{ablation-postprocessing}
\end{table*}

\noindent\textbf{Results}.
Comparing Tables \ref{ablation-preprocessing} and \ref{ablation-postprocessing}, we can notice that performance trend is the same.
However, in the post-processing, the convolutional performance increment is less evident.
Contrary to the original non-local, our version with $7\times7$ kernels obtained a considerably high improvement.

\subsubsection{GRoIE module analysis}

\noindent\textbf{Description}.
Finally, we tested the GRoIE architecture with the best-performing pre- and post-processing modules: a $7\times7$ convolution as pre-processing and non-local with $7 \times 7$ kernels as post-processing.\\

\begin{table*}[bth]
	\begin{center}
		\begin{tabular}{l|c|c|c|c|c|c}
			Method & $AP$ & $AP_{50}$ & $AP_{75}$ & $AP_{s}$ & $AP_{m}$ & $AP_{l}$ \\
			\hline\hline
			baseline          & 37.4 & 58.1 & 40.4 & 21.2 & 41.0 & 48.1 \\
			\one{GRoIE} & \one{39.3} & \one{59.8} & \one{43.0} & \one{23.0} & \one{42.7} & \one{50.8} \\
		\end{tabular}
	\end{center}
	\caption{Best GRoIE configurations.}
	\label{ablation-groie}
\end{table*}

\noindent\textbf{Results}.
From Table \ref{ablation-groie}, we can observe a great improvement in the performance, surpassing the original AP by 1.9\%.

\subsection{Experiments on SBR-CNN}
\label{experiments all}

\begin{table*}[t]
		\begin{tabular}{@{}c|c||c|c|c|c|c|c||c|c|c|c|c|c}
			& & \multicolumn{6}{c||}{Bounding Box} & \multicolumn{6}{c}{Mask} \\
			\# & Method & $AP$ & $AP_{50}$ & $AP_{75}$  & $AP_{s}$ & $AP_{m}$ & $AP_{l}$  & AP & $AP_{50}$ & $AP_{75}$  & $AP_{s}$ & $AP_{m}$ & $AP_{l}$\\
			\hline\hline
			1 & Mask                 & 37.3 & 58.9 & 40.4 & 21.7 & 41.1 & 48.2 & 34.1 & 55.5 & 36.1 & 18.0 & 37.6 & 46.7 \\
			2 & CondInst             & 38.3 & 57.3 & 41.3 & 22.9 & 41.9 & 49.0 & 34.4 & 54.9 & 36.6 & 15.8 & 37.9 & 49.5 \\
			3 & HTC                  & \red{41.7} & \red{60.4} & \red{45.2} & \red{24.0} & \red{44.8} & \red{54.7} & \red{36.9} & \red{57.6} & \red{39.9} & \red{19.8} & \red{39.8} & \red{50.1} \\
			4 & SBR-CNN              & \one{42.0} & \one{61.1} & \one{46.2} & \one{24.2} & \one{45.3} & \one{55.3} & \one{39.2} & \one{58.7} & \one{42.4} & \one{20.6} & \one{42.6} & \one{54.2} \\
			\hline
			5 & GC-Net               & 40.5 & 62.0 & 44.0 & 23.8 & 44.4 & 52.7 & 36.4 & 58.7 & 38.5 & 19.7 & 40.2 & 49.1 \\
			6 & HTC+GC-Net           & \red{43.9} & \red{63.1} & \red{47.7} & \red{26.2} & \red{47.7} & \red{57.6} & \red{38.7} & \red{60.4} & \red{41.7} & \red{21.6} & \red{42.2} & \red{52.5} \\
			7 & SBR-CNN+GC-Net       & \one{44.8} & \one{64.6} & \one{49.0} & \one{27.2} & \one{48.0} & \one{58.8} & \one{41.3} & \one{62.1} & \one{44.7} & \one{23.1} & \one{44.6} & \one{56.4}\\
			\hline
			8 & DCN                  & 41.9 & 62.9 & 45.9 & 24.2 & 45.5 & 55.5 & 37.6 & 60.0 & 40.0 & 20.2 & 40.8 & 51.6 \\
			9 & HTC+DCN              & \red{44.7} & \red{63.8} & \red{48.6} & \red{26.5} & \red{48.2} & \red{60.2} & \red{39.4} & \red{61.2} & \red{42.3} & \red{21.9} & \red{42.7} & \red{54.9} \\
			10 & SBR-CNN+DCN          & \one{45.3} & \one{64.6} & \one{49.7} & \one{27.2} & \one{48.8} & \one{60.6} & \one{41.5} & \one{62.2} & \one{45.0} & \one{22.9} & \one{45.1} & \one{58.0} \\
		\end{tabular}
\caption{Performance of the state-of-the-art models compared with SBR-CNN model. Bold and red values are respectively the best and second-best results.}
\label{sota-coco-bbox-mask-results}
\end{table*}

\noindent\textbf{Description}.
In this experiment, we compare Mark-RCNN, CondInst \cite{tian2020conditional} and HTC with our SBR-CNN (Self-Balanced R-CNN) model with the following configuration: 
the best-performing three-loop model with the rebuilt detection head and MaskIoU head (see row \#4 of Table \ref{tab:compare-MaskIoU-modded}), with our GRoIE having its best configuration (see Table \ref{ablation-groie}) in place of both Bounding Box and Mask RoI extractors.
In addition, we take into account GC-Net \cite{cao2019gcnet} and Deformable Convolutional Networks (DCN) \cite{zhu2019deformable}, investigating whether the performance benefit we bring is independent of the underlying architecture.
To be as fair as possible, we compare also GC-Net and DCN joined with HTC.
For example, HTC+GC-Net means that we considered the combination of both architectures.\\

\noindent\textbf{Results}.
In Table \ref{sota-coco-bbox-mask-results} we see that, independently from the architecture, our SBR-CNN reaches the highest AP values in all metrics in both tasks, even if the counterpart is merged with HTC.
More specifically, fusing other models with SBR-CNN not only maintains the performance increment but also increases the gap in favor of SBR-CNN.

In case of $B_{AP}$, for instance, looking at the $B_{AP}$ in the standalone case (row \#4), SBR-CNN outperforms HTC (row \#3) by a 0.3\% margin only.
But, when combined with GC-Net and DCN, this improvement is even higher (0.9\% in the case of GC-Net - row \#7 vs \#6 - and 0.6\% in the case of DCN - row \#10 vs \#9).
Considering all metrics, the improvement is up by 1.5\% (see $AP_{50}$ in row \#7 vs \#6).

In case of $S_{AP}$, it fluctuates from +2.1\% up to +2.6\%, when comparing SBR-CNN+GC-Net with HTC+GC-Net (row \#7 vs \#6).
Considering all metrics, the highest improvement is +4.1\% (see $AP_l$ in row \#10 vs \#9).
\section{Conclusions and future works}
\label{conclusions}

\begin{figure*}[bth]
    \begin{center}
        \includegraphics[clip, width=0.8\linewidth]{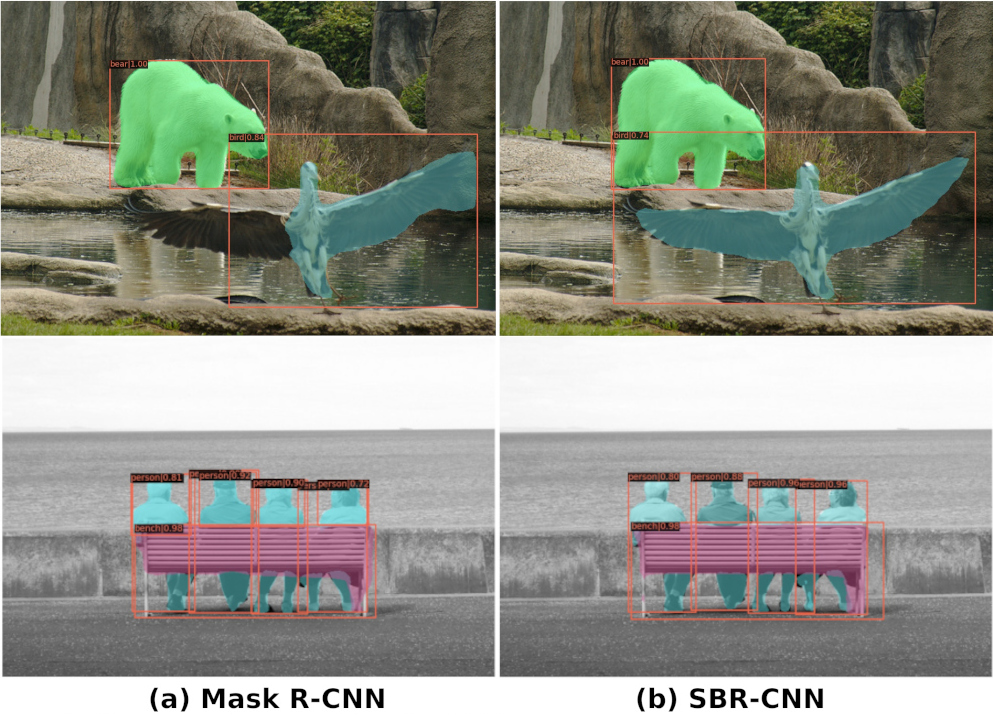}
    \end{center}
    \caption{Examples of instance segmentation comparison between Mask R-CNN (left) and SBR-CNN (right), filtered with a class confidence threshold of 0.7.}
    \label{fig:example-comparison}
\end{figure*}

\begin{figure*}[bth]
    \begin{center}
        \includegraphics[clip, width=0.7\linewidth]{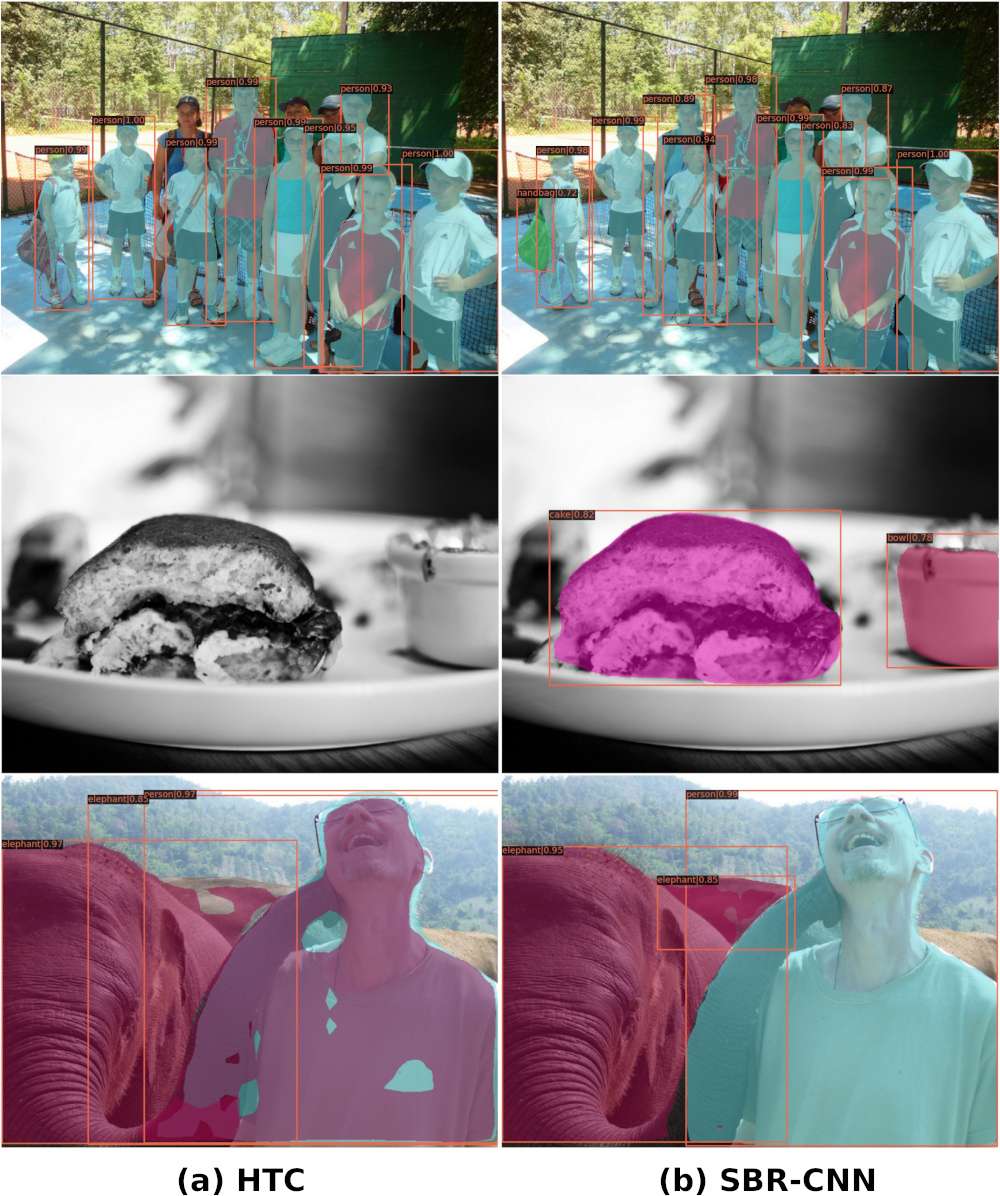}
    \end{center}
    \caption{Examples of instance segmentation comparison between HTC (left) and SBR-CNN (right), filtered with a class confidence threshold of 0.7.}
    \label{fig:example-comparison_htc}
\end{figure*}

We propose a new object detection and instance segmentation architecture called SBR-CNN, which addresses two of intrinsic imbalances which affect two-stage architectures descending from Mask R-CNN:
the IoU Distribution Imbalance of positive input bounding boxes with the help of a new mechanism for refining RoIs through a loop between detection head and RoI extractor, and a loop for mask refinement inside the segmentation head.
Furthermore, we address the Feature Imbalance that afflicts the FPN layers, proposing a better performing RoI Extractor which better integrates low- and high-level information.
Finally, we investigate the effect of a redesign of the model head toward a lightweight fully convolutional solution (FCC).
Our empirical studies confirmed that if the task involves classification, there is the necessity to maintain some spatial sensitivity information by the enhanced non-local block.
Otherwise, when a regression task is involved, a convolutional head is enough.

Our SBR-CNN proves to be successfully integrated into other state-of-the-art models, reaching a 45.3\% AP for object detection and 41.5\% AP for instance segmentation, using only a small backbone such as ResNet-50.
In Figure \ref{fig:example-comparison}, there are some examples of instance segmentation of SBR-CNN compared with a Mask R-CNN.
Many times, our detections are less overconfident and have a more precise segmentation (see the bird on the top).
Our SBR-CNN model also has a tendency to have fewer false positives (see the people on bottom images), maybe as a consequence of less high confidence values.
In Figure \ref{fig:example-comparison_htc} we also compared our model with results obtained by HTC.
Our model could find more objects inside the images, but also for objects found by both network, we can obtains a better segmentation.
The most evident case is the bottommost case.

The SBR-CNN model, formed by the contributions $R^3$-CNN, FCC and GRoIE also carries with it some limitations.
In particular, in the lighter $R^3$-CNN \textit{naive} version, the segmentation head is really effective, making $R^3$-CNN ideal as a replacement for HTC.
The same consideration cannot be made for the detection head.
To compensate for the decrease in performance, it is possible to either use an intermediate version such as the \textit{deeper}, or use \textit{naive} $R^3$-CNN in conjunction with FCC, depending how much is critical the need to decrease as much as possible the number of parameters.
If the second option is chosen, the system has the advantage of making the performance higher and decreasing the size in terms of weights, but with the disadvantage of being much slower on evaluation.
For this reason, in future, it would be advisable to explore equivalent solutions for FCC but which has lower execution times.
Finally, it would be interesting to evaluate in more detail why using two Non-local attention modules, both in GRoIE and in FCC, does not leads to an increase in performance as expected.

\section*{Acknowledgments}
We acknowledge the CINECA award under the ISCRA initiative, for the availability of high performance computing resources and support.

\bibliographystyle{ieeetr}
\bibliography{paper}

\end{document}